\documentclass[10pt,conference]{IEEEtran}
\IEEEoverridecommandlockouts

\usepackage{cite}
\usepackage{amsmath,amssymb,amsfonts}
\usepackage{algorithmic}
\usepackage{graphicx}
\usepackage{textcomp}
\usepackage{xcolor}
\usepackage{hyperref} 
\usepackage{multirow} 
\usepackage{subcaption} 
\usepackage{float}
\usepackage{orcidlink}
\renewcommand{\footnoterule}{\hrule width 0.5\columnwidth \kern 4pt}
\def\BibTeX{{\rm B\kern-.05em{\sc i\kern-.025em b}\kern-.08em
    T\kern-.1667em\lower.7ex\hbox{E}\kern-.125emX}}
\hypersetup{colorlinks=true, linkcolor=black, citecolor=black, urlcolor=blue}

\begin{document}

\title{NERVE: A Neuromorphic Vision and Radar Ensemble for Multi-Sensor Fusion Research}

\author{
\IEEEauthorblockN{Omar Mansour \textsuperscript{*} \orcidlink{0009-0002-4711-848X} \textsuperscript{1}, Pietro Martinello \textsuperscript{*} \orcidlink{0009-0002-4843-7554} \textsuperscript{2}
, Ethan Milon \orcidlink{0009-0005-6590-7353} \textsuperscript{3}, Yingfu Xu \orcidlink{0000-0001-7834-3204} \textsuperscript{4},\\Manolis Sifalakis \orcidlink{0000-0002-0949-2094} \textsuperscript{5}, Guangzhi Tang \textsuperscript{$\dagger$} \orcidlink{0000-0002-0204-9225} \textsuperscript{6}, Amirreza Yousefzadeh \textsuperscript{$\dagger$} \orcidlink{0000-0002-2967-5090}\textsuperscript{1}
\thanks{\textsuperscript{*}Equal contribution.}
\thanks{\textsuperscript{$\dagger$}Equal senior/supervising contribution.}}
\vspace{5pt}
\IEEEauthorblockA{\textsuperscript{1}University of Twente, Netherlands 
\quad
\textsuperscript{2}University of Modena and Reggio Emilia, Italy }
\IEEEauthorblockA{\textsuperscript{3}University of Strasbourg, France \quad \textsuperscript{4}IMEC the Netherlands, Netherlands}
\IEEEauthorblockA{\textsuperscript{5}Innatera, Netherlands \quad \textsuperscript{6}Maastricht University, Netherlands}
\IEEEauthorblockA{Correspondence: omar.mansour3@protonmail.com}
}

\maketitle

\begin{abstract}
We present NERVE (Neuromorphic Vision and Radar Ensemble), a multi-sensor dataset comprising 257 minutes of synchronized recordings from five sensors: two Dynamic Vision Sensors (DVS), an RGB-D camera, and two Radar units (24GHz and 77GHz). Captured across 12 measurement days in office environments, NERVE contains around 600~GB of uncompressed temporally aligned data with around 914,000 frames and around 9.6 million RGB COCO-formatted annotations covering 16 relevant object categories. To evaluate multi-modal fusion, we construct a DVS+Radar subset for human detection and distance estimation. Baseline experiments using feed-forward and recurrent detectors show that combining DVS with 77GHz Radar consistently improves detection, with recurrent models achieving up to 47.5\% mAP and mean absolute Radar distance errors below 1.8~m against LiDAR ground truth. 
\end{abstract}


\section{Introduction}
Event-based cameras, also known as Dynamic Vision Sensors (DVS), capture brightness and contrast changes asynchronously per pixel in the Field of View of the sensor rather than capturing frames at fixed intervals \cite{gallego2020event}. This operating principle produces a continuous stream of events with microsecond temporal resolution, high dynamic range exceeding 120~dB, and potentially reduced power consumption. Importantly, the sparse, asynchronous output of DVS sensors aligns with neuromorphic computing architectures that process information through sparse temporal events rather than dense activations \cite{mead1990neuromorphic}. This compatibility positions DVS as a natural input modality for fast and energy-efficient, event-driven perception systems.

Multi-sensor fusion combines complementary sensor modalities to overcome individual limitations \cite{chandrasekaran2017survey}. Large-scale datasets such as KITTI \cite{geiger2013vision} and nuScenes \cite{caesar2020nuscenes} have advanced fusion research by providing synchronized data from cameras, LiDAR, Radar, and GPS. However, these datasets rely on conventional frame-based sensors and are not compatible with neuromorphic processing pipelines. Conversely, existing neuromorphic datasets focus primarily on single-modality DVS data \cite{detournemire2020large, perot2020learning} or combine DVS with Radar only for specific gesture recognition tasks \cite{muller2023marshalling}. This leaves a gap for a dataset that covers more generic tasks such as scene classification or activity recognition for applications in  smart spaces (indoor, in-cabin, etc.) with data formats suitable for neuromorphic computing research.

DVS-Radar fusion is particularly promising for perception tasks. Radar provides distance measurements and operates reliably under adverse lighting or weather conditions where optical sensors degrade \cite{wei2022mmwave}. DVS contributes high temporal resolution for tracking fast motion. Recent works have successfully applied this sensor combination to SLAM \cite{safa2023fusing} and drone landing localization \cite{wang2025ultra}, demonstrating the modalities' complementary nature for navigation tasks. However, these methods-focused works do not provide datasets for general object detection across multiple categories. To enable perception research with this sensor combination, we present NERVE (Neuromorphic Vision and Radar Ensemble), a multi-sensor dataset that pairs DVS cameras with mmWave Radar in indoor environments.

\begin{figure*}[t]
\centering
\begin{subfigure}[t]{0.48\textwidth}
    \centering
    \includegraphics[width=\linewidth]{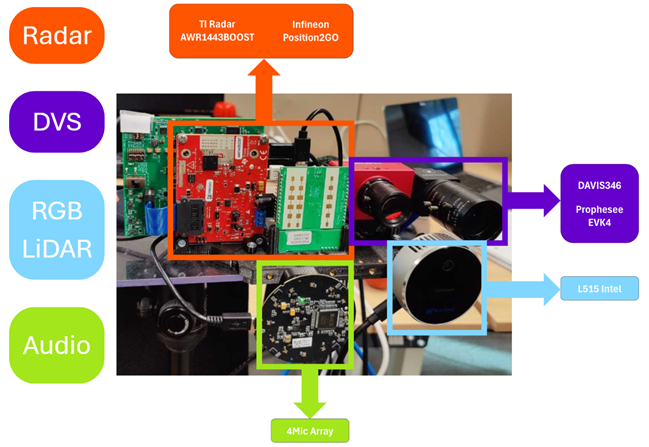}
    \caption{NERVE sensor setup. Audio and RGB data are not published due to privacy concerns, but were used for ground truth labeling and validation.}
    \label{fig:Acquisition_Setup}
\end{subfigure}\hfill
\begin{subfigure}[t]{0.48\textwidth}
    \centering
    \includegraphics[width=\linewidth]{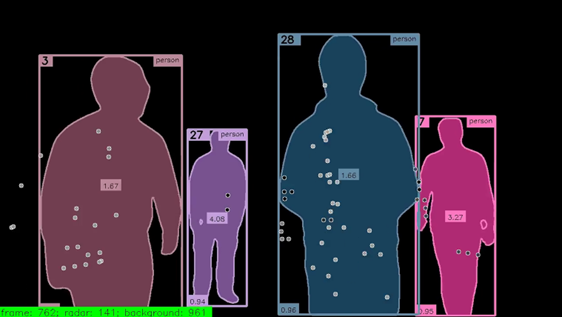}
    \caption{Overlay of annotations on an early fusion frame that combines Radar point clouds and stacked DVS events.}
    \label{fig:Picture1}
\end{subfigure}
\caption{Overview of the NERVE dataset: (a) the multi-sensor acquisition setup; (b) an example fused frame with annotations.}
\label{fig:setup_and_overlay}
\end{figure*}

Our contributions are:
\begin{itemize}
    \item A temporally aligned multi-sensor dataset combining neuromorphic vision, Radar, and depth sensing with semi-automated dense annotations designed for both conventional and neuromorphic processing (Fig.\ref{fig:Acquisition_Setup}), released following FAIR principles~\cite{wilkinson2016fair} to ensure findability, accessibility, interoperability, and reusability.
    \item An open-source software toolkit providing preprocessing pipelines and format conversion utilities that transform raw sensor data into training-ready formats compatible with detection frameworks such as YOLO ((Fig.\ref{fig:Picture1}).
    \item Baseline experiments demonstrating that DVS-Radar fusion improves human detection over single-modality approaches.
\end{itemize}

The dataset, preprocessing toolkit, and baseline code is publically available at \url{https://github.com/nerveproject-ut/nerve}.

\section{Related Work}

\textbf{Event-Based Vision Datasets.}
Early event-based datasets were generated by converting frame-based image collections. Orchard et al.~\cite{orchard2015converting} produced N-MNIST and N-Caltech101 by moving a DVS camera in front of displayed images, enabling initial classification research on neuromorphic hardware. However, these converted datasets lack the complexity of real-world motion patterns and scene dynamics. Event simulators such as ESIM~\cite{rebecq2018esim} and video-to-event conversion methods~\cite{gehrig2020video} provide scalable alternatives for generating training data, though synthetic events may not accurately capture sensor-specific noise characteristics or the large dynamic range and non-uniform temporal event distributions of real DVS hardware.

Real-world DVS datasets have grown substantially in scale and application scope. The DDD17~\cite{binas2017ddd17} and DDD20~\cite{hu2020ddd20} driving datasets provide end-to-end recordings for steering prediction, demonstrating DVS utility in automotive control albeit without object-level annotations. For detection tasks, the GEN1 dataset~\cite{detournemire2020large} offers 39 hours of automotive recordings with 255,000 manually labeled bounding boxes across two categories (pedestrians and vehicles), while the 1~Mpx dataset~\cite{perot2020learning} provides higher spatial resolution (1280$\times$720) with automatically generated labels. Both datasets capture scenes from dashboard-mounted cameras at relatively fixed distances, resulting in objects concentrated along the horizon line. Surveillance-focused datasets~\cite{miao2019neuromorphic} target pedestrian detection and fall detection using stationary cameras, limiting viewpoint diversity. PEDRo~\cite{boretti2023pedro} addresses robotics scenarios with a moving DAVIS346 camera across diverse indoor and outdoor environments, demonstrating improved generalization to egocentric viewpoints. Despite this progress, all existing DVS detection datasets provide single-modality event data without complementary sensors that could supply depth, velocity, or environmental robustness.\\

\textbf{Multi-Sensor Fusion Datasets.}
Autonomous driving research has benefited substantially from large-scale multi-sensor benchmarks. The KITTI dataset~\cite{geiger2013vision} combines high-resolution RGB stereo cameras, a Velodyne 64-beam LiDAR, and GPS/IMU, establishing standardized evaluation protocols for stereo matching, optical flow, and 3D object detection. nuScenes~\cite{caesar2020nuscenes} extends multi-sensor coverage with six cameras providing 360-degree visibility, five Radar units, one LiDAR sensor, and GPS/IMU, annotating over 1.4 million images with 3D bounding boxes across 23 object categories in 1000 diverse urban scenes. The MVSEC dataset~\cite{zhu2018multivehicle} introduces stereo event cameras alongside conventional frame-based sensors for 3D perception research, providing synchronized DVS, grayscale frames, IMU, and LiDAR data. These datasets have driven significant advances in perception algorithms but none combine event-based neuromorphic sensors with Radar for object detection, leaving this complementary sensor pairing largely unexplored despite the potential for improved robustness in challenging conditions.\\

\textbf{Neuromorphic Multi-Modal Sensing.}
The Marshalling Signals dataset~\cite{muller2023marshalling} represents the primary existing work combining DVS with Radar, pairing an 8GHz UWB FMCW Radar with a DVS camera for aircraft marshalling gesture recognition. The authors propose sparse binary encoding of Radar signals that reduces data rates while preserving frequency content suitable for low-power neuromorphic processing. However, the dataset focuses on a constrained recognition task (10 predefined full-body gestures) rather than general object detection, and the Single-Input Single-Output (SISO) Radar does not have angular coverage which is needed for most radar system applications. While mmWave Radar and vision fusion has been surveyed for automotive contexts~\cite{wei2022mmwave}, existing work focuses exclusively on frame-based cameras rather than neuromorphic sensors. Consequently, DVS-Radar fusion remains an open research direction lacking both extensive standardized datasets and established baselines.\\ \\

NERVE aims to fill this gap by providing a multi-sensor dataset with dense annotations specifically designed for perception research in service robotics scenarios, complementing the algorithmic advances in DVS-Radar fusion with a reusable benchmark for the community. 

\section{The NERVE Dataset}
NERVE targets human detection in smart office scenarios, where sensors observe people and indoor objects at close range (0.5--6~m) from unrestricted orientations. Unlike automotive datasets with dashboard-mounted cameras focused on highway distances, this application context requires diverse viewpoints and proximity-based interactions.

The dataset comprises 117 recording sessions organized into 13 session groups, captured across 12 measurement days in office environments. Sessions within the same group share identical conditions (environment, lighting, participants), while different groups provide diversity.

\subsection{Acquisition Setup}
All sensors are mounted on a 3D-printed platform attached to a tripod (Figure~\ref{fig:Acquisition_Setup}), maintaining rigid geometric relationships throughout data collection. The platform accommodates five sensors, two DVS cameras providing complementary spatial resolutions, one RGB-D sensor for depth ground truth and automatic labeling, and two Radar units operating at different frequencies.

We provide redundancy for both vision and Radar sensors (a high-resolution set and a low-resolution set), allowing researchers to explore trade-offs among accuracy, algorithmic complexity, and cost when building a smart office IoT sensor set. For vision, the Prophesee EVK4 captures events at 1280$\times$720 with 120~dB dynamic range, while the IniVation DAVIS346 offers a lower resolution of 346$\times$260 with the same dynamic range. For Radar, the TI AWR1443BOOST operates at 77~GHz with 1536~MHz bandwidth and 9.77~cm range resolution across 3~TX/4~RX channels, while the Infineon Position2GO operates at 24~GHz with 200~MHz bandwidth and 0.9~m range resolution. The Intel RealSense L515 combines a Full-HD RGB camera (1280$\times$720 at 60~fps) with a LiDAR-based depth sensor (1024$\times$768 at 30~fps, range 0.25--9.0~m). The depth stream provides per-pixel distance measurements with manufacturer-specified accuracy of $\pm$5~mm at 1~m~\cite{intel2020l515}, serving as ground truth for Radar distance validation.

\subsection{Time Alignment and Calibration}
Multi-sensor fusion requires precise temporal and spatial alignment. We employ a hybrid synchronization approach that combines hardware signaling for sensor pairs with stringent timing requirements and software timestamps for the remaining sensors.

The DAVIS346 and Position2GO Radar achieve hardware-level synchronization following~\cite{muller2023marshalling}, where the Radar's pulse repetition interval (PRI) signal is fed directly into the DVS event stream as a rising-edge trigger. This enables chirp-level alignment between Radar frames and DVS events. The remaining sensors (EVK4, L515, TI Radar) are synchronized via operating system timestamps recorded at acquisition start. Empirical evaluation confirms synchronization accuracy within $\pm$20~ms across all modalities.

For spatial calibration, we used ChArUco board patterns~\cite{an2018charuco} for optical sensors. Unlike standard checkerboards, ChArUco patterns tolerate partial occlusions, enabling robust corner detection at field-of-view boundaries. For DVS cameras, we reconstruct video frames from accumulated events while slowly moving the calibration pattern, then apply standard calibration techniques to the reconstructed frames. The Intel L515 serves as the common reference frame; extrinsic parameters transform annotations between sensor coordinate systems. Radar-to-camera alignment uses the L515 depth stream to identify corresponding 3D points.

\subsection{Annotation Protocol}
\label{sec:annotation}
Manually generating dense annotations for around 914,000 frames across multiple sensors would require prohibitive effort. We adopt an automatic labeling pipeline that processes RGB video through pre-trained models, then maps the resulting annotations to other sensor coordinate systems.\\

\textbf{Label Generation.} RGB frames from the Intel L515 are processed using YOLOv8x~\cite{jocher2023yolov8} for object detection, instance segmentation, and pose estimation. For each detected object, we extract (1) bounding box coordinates, (2) category label from 16 classes, (3) instance segmentation mask, (4) skeleton keypoints for human detections (17 COCO keypoints), and (5) temporary entity ID enabling frame-to-frame tracking. The depth stream provides per-bounding-box distance estimates by sampling high-confidence depth pixels within each segmentation mask.\\

\begin{figure}[H]
    \centering
    \includegraphics[width=\columnwidth]{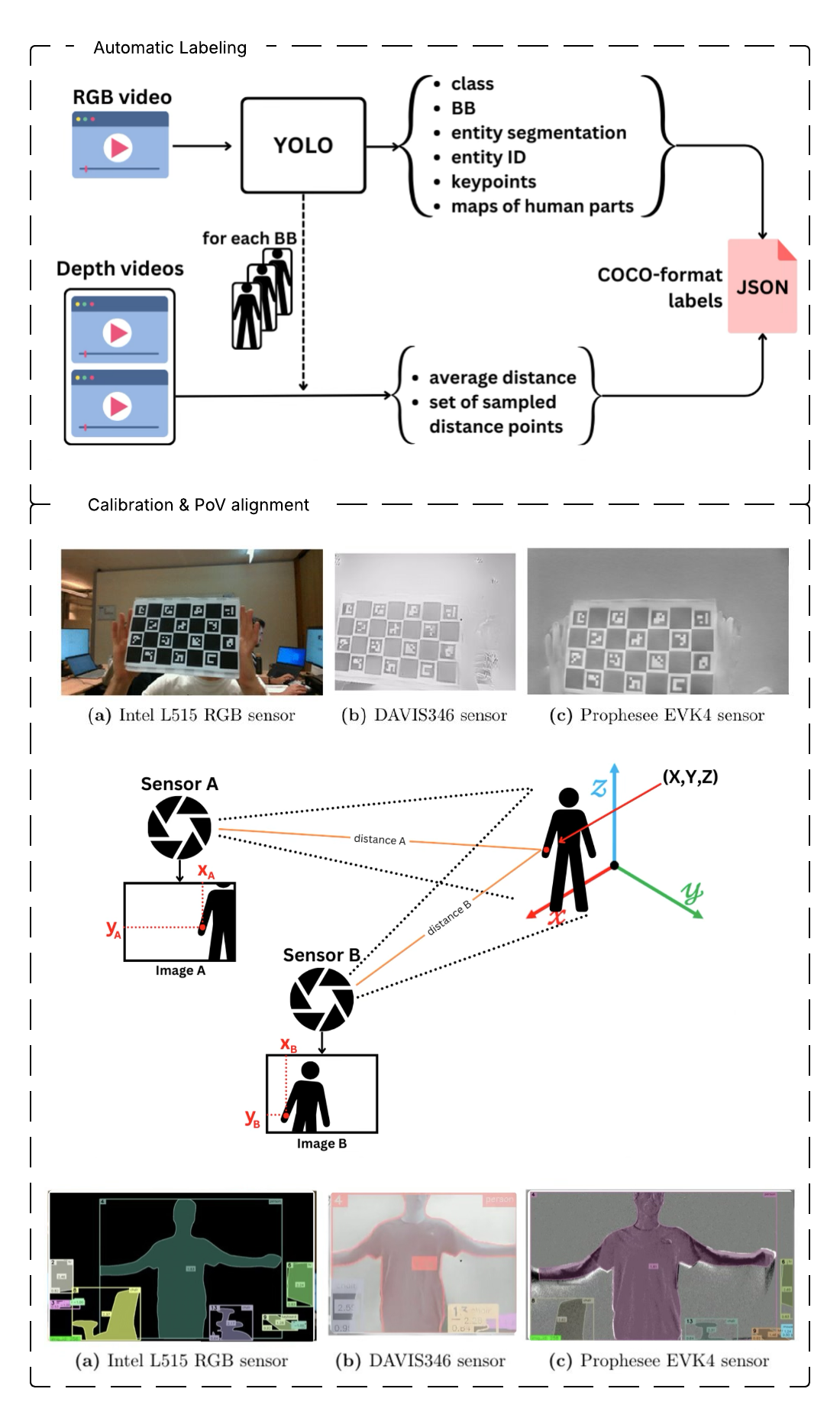}
    \caption{Block diagram of the automatic annotation pipeline: RGB frames are processed by YOLOv8x for detection, segmentation, and pose estimation, then mapped to DVS coordinate systems via extrinsic calibration and depth-based 3D reprojection.}
    \label{fig:annotation_pipeline}
\end{figure}

\textbf{Cross-Sensor Mapping.} Annotations generated in RGB space are transformed to DVS coordinate systems using extrinsic calibration parameters and the L515 depth stream. For each annotated pixel $(x_{rgb}, y_{rgb})$, we retrieve its 3D position from the corresponding depth value, transform it to the target sensor frame, and project it onto the target image plane. This process accounts for lens distortion in both source and target sensors. Fig.~\ref{fig:annotation_pipeline} illustrates the complete automatic labeling and cross-sensor alignment pipeline.\\

\textbf{Privacy Preservation.} Raw RGB video and audio recordings from the ReSpeaker 4-mic array (visible in Figure~\ref{fig:Acquisition_Setup}) are excluded from the released dataset to protect participant privacy. All semantic information (object locations, categories, poses) is preserved in the annotations, enabling full reproducibility of detection experiments without exposing identifiable imagery or voice data.\\

\subsection{Data Format}
Each recording session produces a directory containing synchronized data from all sensors in standardized formats:

\begin{itemize}
    \item \textbf{DVS Events:} Prophesee EVK4 events stored in HDF5 format; DAVIS346 events in custom RAD format.
    \item \textbf{Radar Data:} TI AWR1443BOOST provides raw ADC frames in HDF5 with full metadata, enabling custom signal processing. Position2GO data is embedded in the DAVIS346 RAD files.
    \item \textbf{Annotations:} COCO-format JSON files containing bounding boxes, segmentation masks, keypoints, and distance estimates, provided in both RGB and DVS coordinate spaces.
\end{itemize}

\subsection{Software Tools and FAIR Compliance}
To maximize research utility, NERVE is released with an accompanying open-source software toolkit. The toolkit provides (1) preprocessing pipelines that convert raw sensor streams into normalized formats for all sensors except radar, (2) event-to-frame conversion utilities supporting configurable temporal windows and accumulation strategies, (3) format converters that generate training-ready datasets compatible with detection frameworks that f.e. use YOLO, and (4) visualization utilities for multi-modal data inspection. All code is publicly available alongside the dataset.

The dataset release adheres to the FAIR principles~\cite{wilkinson2016fair}. \textit{Findability} is ensured through persistent identifiers and comprehensive metadata describing sensor configurations, recording conditions, and annotation schemas. \textit{Accessibility} is provided via standard protocols with clear licensing terms. \textit{Interoperability} is achieved through standardized file formats (HDF5, COCO JSON, MP4) and documented coordinate transformations between sensors. \textit{Reusability} is supported by detailed documentation and the accompanying software toolkit.

\subsection{Statistics and Analysis}
The dataset totals 257 minutes of recordings across 117 sessions (13 session groups, 12 measurement days), occupying approximately 600~GB of storage. From around 914,000 annotated frames we obtain around 9.6 million annotations spanning 16 object categories, averaging 10.4 objects per frame. The annotation distribution reflects typical office scenes: \textit{chair} (3.5M) and \textit{person} (2.2M) dominate, followed by \textit{laptop} (1.3M) and \textit{tv/monitor} (500K), while infrequent categories such as \textit{book} (40K) and \textit{apple} (5.5K) capture rarer office interactions.

\begin{figure*}[t]
\centering
\begin{subfigure}[t]{0.32\textwidth}
    \centering
    \includegraphics[width=\linewidth]{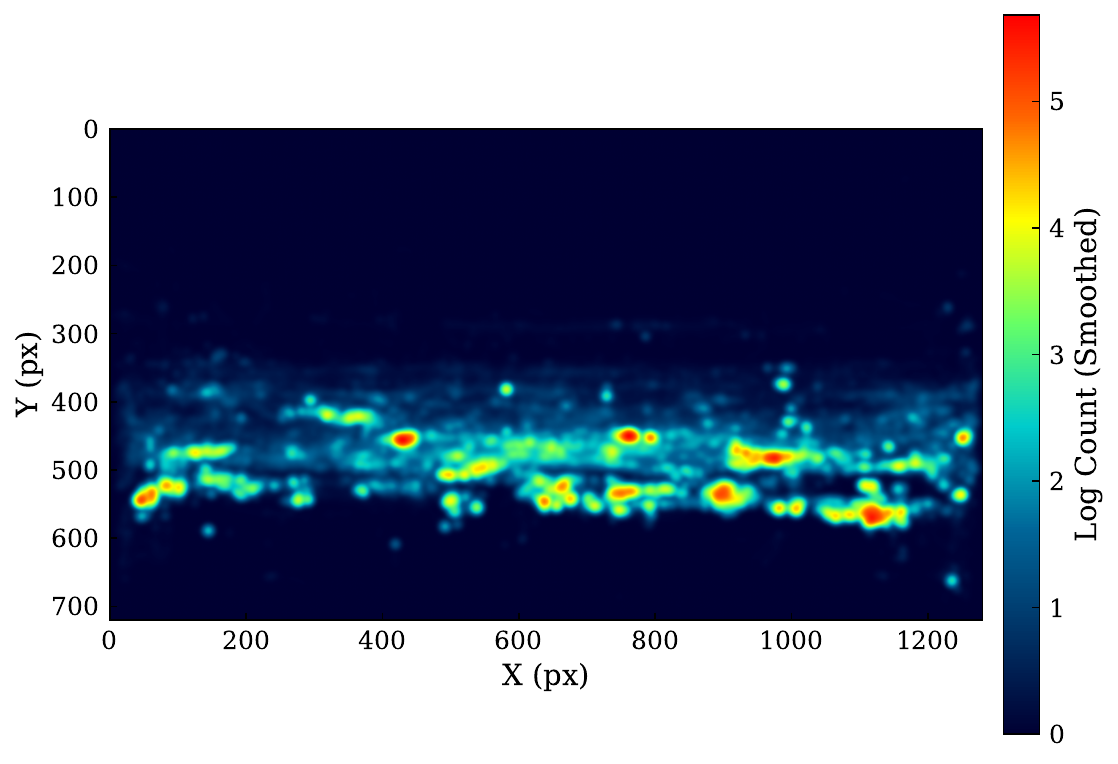}
    \caption{Bounding box center heatmap.}
    \label{fig:heatmap}
\end{subfigure}\hfill
\begin{subfigure}[t]{0.32\textwidth}
    \centering
    \includegraphics[width=\linewidth]{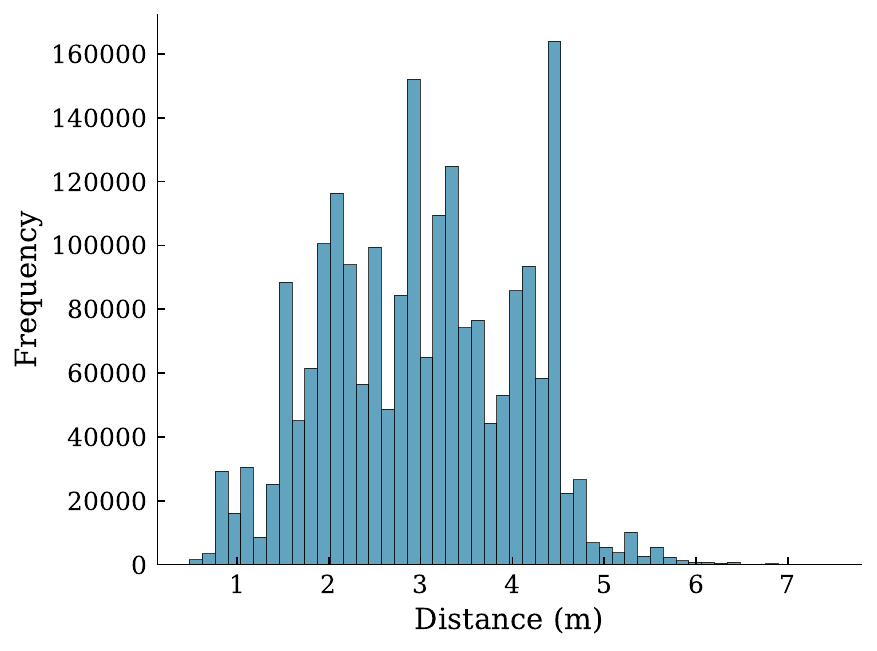}
    \caption{Ground truth distances.}
    \label{fig:groundtruthdistances}
\end{subfigure}\hfill
\begin{subfigure}[t]{0.32\textwidth}
    \centering
    \includegraphics[width=\linewidth]{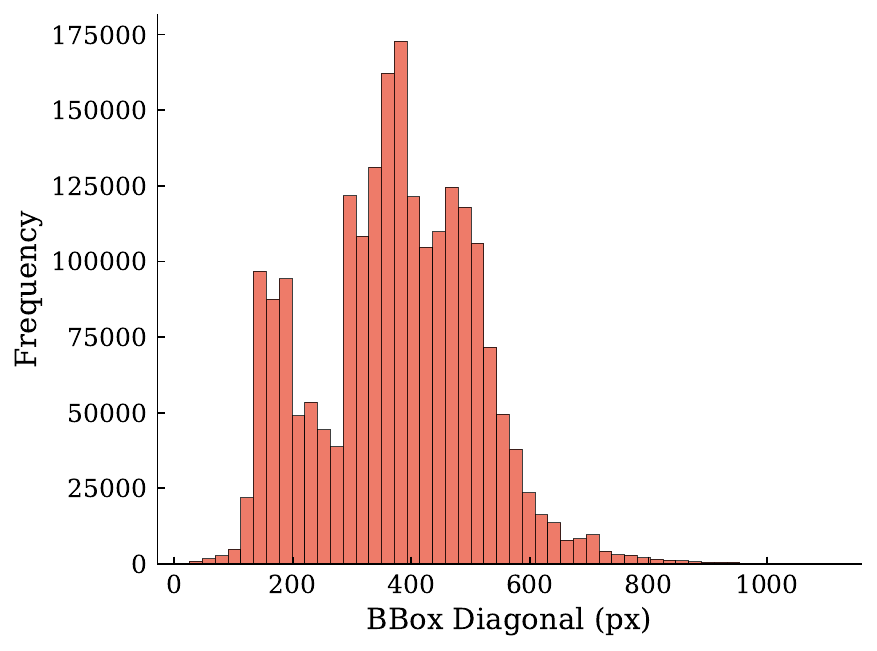}
    \caption{Ground truth bounding-box diagonals.}
    \label{fig:groundtruthbb}
\end{subfigure}
\caption{NERVE dataset distributions: (a) spatial distribution of bounding box centers for class \textit{human}, showing diverse spatial coverage; (b) distance distribution measured by Intel L515; (c) bounding-box diagonal distribution from the annotation protocol.}
\label{fig:groundtruths_combined}
\end{figure*}

Fig.~\ref{fig:heatmap} shows the spatial distribution of bounding box centers that form a uniform coverage across the field of view, reflecting the varied viewpoints encountered in indoor environments and interactions with nearby humans. Additionally, Fig.~\ref{fig:groundtruthdistances} and Fig.~\ref{fig:groundtruthbb} show respectively the distribution of ground truth distances measured using the Intel L515 LiDAR Camera and the distribution of ground truth bounding box diagonals as generated by the Annotation Protocol described in ~\ref{sec:annotation}.

Compared with existing DVS and multi-sensor datasets, NERVE is the first to combine neuromorphic vision, Radar, and depth sensing in a single benchmark. Prior DVS datasets such as GEN1~\cite{detournemire2020large} , 1Mpx~\cite{perot2020learning} , and PEDRo~\cite{boretti2023pedro} provide event-only data without complementary modalities. MVSEC~\cite{zhu2018multivehicle} adds depth but lacks Radar, and the Marshalling dataset~\cite{muller2023marshalling} pairs DVS with Radar yet targets only gesture recognition (10 classes) without depth. 

\section{Experiments}
We evaluate DVS-Radar fusion for human detection and distance estimation using a subset of NERVE. These experiments establish baseline performance for multi-modal neuromorphic perception and validate the utility of Radar data for improving event-based object detection.

\subsection{DVS-Radar Subset Construction}
From the full NERVE dataset, we construct two DVS-Radar subsets: a low-resolution subset pairing the IniVation DAVIS346 with the 77~GHz TI Radar, and a high-resolution subset pairing the Prophesee EVK4 with the same Radar. The two modalities are complementary, with the DVS camera providing spatial information through asynchronous event outputs, while the TI Radar supplies precise distance measurements at a 9.77~cm range resolution.

\textbf{Event Representation.}\label{sec:event_representation} Raw DVS events are converted to dense tensor representations using one of the multiple preprocessing algorithms~\cite{gehrig2023rvt,silva2024reyolov8} available in the public codebase. \textit{Stacked Histograms (SHist)} count events within $B$ temporal bins per polarity, yielding $2B \times H \times W$ tensors; \textit{Voxel Grid} applies bilinear temporal interpolation with polarity-based weighting for smoother distributions; \textit{Mixed-Density Event Stacks (MDES)} use non-uniform temporal bins that prioritize recent events by sampling only the last event per bin; and \textit{Volume of Ternary Event Images (VTEI)} encode events as ternary values ($+1$, $-1$, $0$) for memory-efficient processing. All representations use a 16.67~ms accumulation window, enabling compatibility with standard convolutional architectures while preserving temporal dynamics.\\

\textbf{Radar Processing.} Raw ADC samples from the TI Radar undergo 2D FFT-based range-Doppler processing, followed by CFAR (Constant False Alarm Rate) peak detection to generate sparse point clouds encoding range, velocity, and intensity per target. These points are projected into the DVS image space using extrinsic calibration parameters, yielding a sparse spatial map where non-zero pixels indicate detected targets. Since the Radar operates at 11.1~fps (90~ms frame period) while DVS can generate representations at arbitrary rates, the fused subset is aligned to the Radar frame rate: each Radar frame is paired with the temporally nearest DVS histogram, resulting in synchronized frame pairs at 11.1~fps. Due to the usage of proprietary code during the radar processing, the exact implementation details are not provided and will not be included in the publicly available codebase instead the dataset contains already processed radar data.\\

\textbf{Dataset Split.} The subset is divided 80-10-10 for training, validation, and testing. Critically, sessions from the same recording group are never split across training and test or validation partitions, preventing data leakage from similar environmental conditions or participants appearing in both training and test sets. The list of sessions belonging to each split will be contained inside of the public codebase.

\begin{table*}[!t]
\caption{Human Detection and Distance Estimation Results}
\begin{center}
\footnotesize
\renewcommand{\arraystretch}{1.3}
\setlength{\tabcolsep}{5pt}
\resizebox{\textwidth}{!}{
\begin{tabular}{|l|cccc|cccc|cc|}
\hline
& \multicolumn{4}{c|}{\textit{Feed-forward (scratch)}} & \multicolumn{4}{c|}{\textit{Recurrent (scratch)}} & \multicolumn{2}{c|}{\textit{Zero-shot}} \\
& \multicolumn{2}{c}{YOLOv8-nano} & \multicolumn{2}{c|}{YoloX-Tiny} & \multicolumn{2}{c}{RVT-Tiny} & \multicolumn{2}{c|}{ReYOLOv8-nano} & RVT-Tiny & ReYOLOv8-n \\
& Low & High & Low & High & Low & High & Low & High & (Gen1)$^\dagger$ & (PEDRo)$^\dagger$ \\
\hline
mAP$_{50:95}$ (DVS) & 21.8\% & 24.6\% & 20.3\% & 21.5\% & 40.1\% & 45.3\% & 30.4\% & 32.7\% & 10.1\% & 20.1\% \\
mAP$_{50:95}$ (DVS+Radar) & 27.9\% & 28.0\% & 24.5\% & 25.2\% & 39.4\% & 47.5\% & 32.3\% & 34.8\% & --- & --- \\
MAE (m) & 0.58 & 0.59 & 0.89 & 0.64 & 1.80 & 1.33 & 0.73 & 0.61 & --- & --- \\
\hline
\multicolumn{11}{l}{\scriptsize Low = low-res subset (DAVIS346, 346$\times$260), High = high-res subset (EVK4, 1280$\times$720) \qquad \qquad \qquad \qquad \qquad \quad $^\dagger$Pretrained checkpoint source} \\
\end{tabular}}
\label{tab:results}
\end{center}
\end{table*}

\subsection{Model Architecture and Training}
We evaluate four detector architectures spanning feed-forward and recurrent paradigms. For feed-forward baselines, we employ YOLOv8-nano~\cite{jocher2023yolov8} and YoloX-nano~\cite{ge2021yolox}, lightweight single-stage detectors suitable for edge deployment. For recurrent models that exploit temporal context, we evaluate RVT-tiny~\cite{gehrig2023rvt}, a Recurrent Vision Transformer designed specifically for event-based detection, and ReYOLOv8-nano~\cite{silva2024reyolov8}, a recurrent adaptation of YOLOv8.

\textbf{Architecture Modifications.} For all models, the classification head is removed for single-class (person) detection, retaining only the localization branch that predicts bounding box coordinates and objectness confidence. For fusion experiments, the Radar point cloud projection is added as an additional input channel, providing the network with explicit spatial cues about object distances. 

\textbf{Training Protocol.} We evaluate two initialization strategies: (1) training from scratch and (2) zero-shot transfer using pretrained weights from event-based datasets (PEDRo~\cite{boretti2023pedro} for ReYOLOv8, Gen1~\cite{detournemire2020large} for RVT). Training uses standard detection losses combining L1 localization loss, binary cross-entropy confidence loss, and IoU loss. For event representation, we use the best-performing algorithm from those described in Section~\ref{sec:event_representation}, configured with 5 temporal bins yielding a $5 \times H \times W$ input tensor; for fusion experiments, the Radar projection is concatenated as an additional channel ($6 \times H \times W$). For recurrent models (RVT-tiny, ReYOLOv8-nano), the dataset is processed into temporal clips of 5-10 frames with a stride equivalent to the number of clip frames, enabling the models to leverage motion history for improved detection of stationary subjects.

\textbf{Distance Estimation.} We add a dedicated distance classification head that operates in parallel with the detection head. Rather than directly regressing distance values, the head predicts a probability distribution over discrete distance bins (100 bins spanning 0–10 meters). For each detection anchor, the distance branch applies two 3×3 convolutions, followed by a 1×1 convolution that outputs bin logits. Training uses Binary Cross-Entropy loss against one-hot-encoded ground-truth distances derived from Radar measurements. In inference, the predicted distance is obtained by taking the argmax of the sigmoid-activated bin logits and converting the bin index back to meters.

\subsection{Evaluation Metrics}
Following standard practice in object detection literature, we evaluate using COCO-standard Average Precision (AP) computed across IoU thresholds from 0.50 to 0.95. For distance estimation, we report Mean Absolute Error (MAE) against LiDAR ground truth.

\subsection{Detection Results}
Table~\ref{tab:results} compares detection and distance estimation performance across all evaluated architectures. VTEI consistently yields the best results for recurrent models, while feed-forward networks benefit from SHIST representations. Recurrent architectures substantially outperform feed-forward models, with RVT-Tiny achieving the highest mAP of 47.5\% on the high resolution subset when fused with Radar. Radar fusion improves detection for most configurations, with absolute mAP gains up to 6 percentage points (YOLOv8-nano). The high resolution subset consistently outperforms the low resolution subset across all recurrent models, reflecting the benefits of higher spatial resolution (1280$\times$720 vs 346$\times$260) for detection tasks. This presents a certain tradeoff for IoT where reduced data volume and potential power savings using the low resolution subset has to be weighed against the performance gain experienced by the high resolution subset.

\subsection{Discussion}
\textbf{Sensor Fusion Benefits.} Adding Radar improves detection for most model-sensor combinations, with feed-forward models showing the largest relative gains. However, fusion benefits are architecture-dependent: RVT-Tiny on DAVIS shows a slight decrease with Radar, while the same architecture on Prophesee improves. Running multiple trial runs and determining the statistical significance of this difference yields that the margin for the RVT-Tiny model trained on the low resolution subset falls within statistical error.

\textbf{Temporal Modeling.} Recurrent architectures substantially outperform feed-forward models, with RVT-Tiny (Prophesee) achieving 45.3\% mAP on DVS-only input compared to 24.6\% for YOLOv8-nano, an 84\% relative improvement. This demonstrates that temporal context is critical for motion-dependent sensing, enabling detection of subjects during brief stationary periods by leveraging recent motion history. RVT-Tiny consistently outperforms ReYOLOv8-nano across both sensor configurations, indicating that transformer-based temporal modeling is particularly effective for event data.

\textbf{Distance Estimation Trade-offs.} YOLOv8-nano achieves the lowest distance MAE (0.58~m) among feed-forward models, while recurrent models show mixed distance errors ranging from 0.61~m (ReYOLOv8-nano High) to 1.80~m (RVT-Tiny Low) despite superior detection performance. This suggests a trade-off between detection coverage and distance accuracy: recurrent models detect more objects at challenging ranges where distance estimation is less reliable, warranting further investigation into joint optimization strategies.

\section{Conclusion}
Existing neuromorphic vision datasets lack the multi-sensor coverage needed for robust perception research, while multi-sensor fusion benchmarks rely on frame-based cameras, which are not very suitable with spike-based processing. We presented NERVE, a dataset that bridges this gap by combining two DVS cameras, two Radar units (24~GHz and 77~GHz), and RGB-D sensing with around 9.6 million COCO-format annotations across 16 object categories. The 257 minutes of time aligned recordings in office environments, with automatic cross-sensor label mapping, provide a foundation for neuromorphic multimodal perception research. Following FAIR principles, the dataset is released with comprehensive metadata, standardized formats, and an open-source software toolkit that provides preprocessing pipelines and format converters for seamless integration with detection frameworks.

Baseline experiments across four architectures demonstrate benefits of DVS-Radar fusion for most configurations, with mAP improvements up to 6 percentage points for feed-forward models. Recurrent models (RVT-Tiny, ReYOLOv8-nano) substantially outperform feed-forward approaches, with RVT-Tiny achieving 47.5\% mAP on the Prophesee-Radar subset, confirming that temporal context is essential for motion-dependent sensing. The high resolution subset consistently outperforms the low resolution subset, highlighting the value of higher-resolution event data for detection tasks. \\

\textbf{Limitations.} NERVE is constrained to indoor office environments with a stationary sensor platform, limiting generalization to outdoor scenarios. Automatic labeling via YOLOv8 inherits the detector's failure modes, and the 16 object categories reflect COCO classes detectable in office settings rather than comprehensive indoor object coverage. The DVS-Radar subset evaluates only one sensor combination; the full dataset supports additional pairings (e.g., DAVIS346 with Position2GO Radar) that remain unexplored.\\

\textbf{Future Work.} NERVE enables several research directions, such as (1) SNN architectures that exploit temporal dynamics of asynchronous DVS and Radar data or (2) advanced temporal models further addressing the static-subject challenge.\\

\section*{Acknowledgment}

This work is financially supported by DAIS (KDT JU grant agreement No. 101007273) and the FAIR Data Fund (4TU community fund, Netherlands). AI tools have been used to edit text paragraphs across all sections and correctly format figures and tables.

\bibliographystyle{IEEEtran}
\bibliography{references}

\end{document}